\documentclass[conference]{IEEEtran}
\IEEEoverridecommandlockouts
\usepackage{flushend,cite}
\usepackage{amsmath,amssymb,amsfonts}
\usepackage{algorithmic}
\usepackage{graphicx}
\usepackage{textcomp}
\usepackage{xcolor}
\usepackage{booktabs}
\usepackage{subfigure}
\usepackage[hyphens]{url}
\usepackage{hyperref}
\usepackage{booktabs}

\usepackage{placeins}
\usepackage{float}
\restylefloat{figure}

\usepackage{sidecap}

\usepackage[hyphens]{url}
\usepackage{hyperref}
\usepackage{booktabs}
\usepackage{tikz}
\usetikzlibrary{positioning}
\usetikzlibrary{shapes.geometric}

\usepackage{balance}

\def\BibTeX{{\rm B\kern-.05em{\sc i\kern-.025em b}\kern-.08em
    T\kern-.1667em\lower.7ex\hbox{E}\kern-.125emX}}
\begin{document}

\title{Integration of Sentinel-1 and Sentinel-2
data for Earth surface classification using Machine Learning
algorithms implemented on  Google Earth Engine}
\makeatletter
\newcommand{\newlineauthors}{%
  \end{@IEEEauthorhalign}\hfill\mbox{}\par
  \mbox{}\hfill\begin{@IEEEauthorhalign}
}
\makeatother

\author{\IEEEauthorblockN{1\textsuperscript{st} Francesca Razzano}
\IEEEauthorblockA{\textit{Engineering Department} \\
\textit{University of Sannio}\\
Benevento, Italy  \\
f.razzano3$@$studenti.unisannio.it}
\and
\IEEEauthorblockN{2\textsuperscript{nd} Mariapia Rita Iandolo}
\IEEEauthorblockA{\textit{Engineering Department} \\
\textit{University of Sannio}\\
Benevento, Italy \\
m.iandolo1$@$studenti.unisannio.it}
\and
\IEEEauthorblockN{3\textsuperscript{rd} Chiara Zarro}
\IEEEauthorblockA{\textit{Aerospace Department} \\
\textit{Intelligentia srl}\\
Benevento, Italy \\ chiara.zarro@intelligentia.it
}
\and
\newlineauthors

\IEEEauthorblockN{4\textsuperscript{th} G. S. Yogesh}
\IEEEauthorblockA{\textit{Engineering Department} \\
\textit{East Point College}\\
Bangalore, India  \\
gs.yogesh@eastpoint.ac.in}
\and
\IEEEauthorblockN{5\textsuperscript{th} Silvia Liberata Ullo}
\IEEEauthorblockA{\textit{Engineering Department} \\
\textit{University of Sannio}\\
Benevento, Italy \\
ullo@unisannio.it}
}

\maketitle

\begin{abstract}
In this study,  Synthetic Aperture Radar (SAR) and optical data are both considered for Earth surface classification. Specifically, the  integration of Sentinel-1 (S-1) and Sentinel-2 (S-2)  data is carried out through supervised Machine Learning (ML) algorithms implemented on the Google Earth Engine (GEE) platform for the classification of a particular region of interest. 
Achieved results demonstrate how in this case radar and optical remote detection provide complementary information, benefiting surface cover classification and generally leading to increased mapping accuracy. In addition, this paper works in the direction of proving the emerging role of GEE as an effective cloud-based tool for handling large amounts of satellite data. 


\end{abstract}

\begin{IEEEkeywords}
Radar and optical data integration, Sentinel-1, Sentinel-2, Earth surface classification, Machine Learning algorithms, Google Earth Engine platform
\end{IEEEkeywords}

\section{Introduction}
In many areas of remote sensing, the research process often involves multimodal input data, since due to the rich characteristics of natural phenomena, it is rare for a single mode to provide complete knowledge of the observed phenomenon \cite{KHALEGHI201328, 7214350, KHALEGHI2013282013}. 
The real purpose of multimodality is to extract and combine important information from individual sensors and use this mixed functionality to solve a given problem. 
Thus, the expected output will have a representation in terms of results richer than the individual modes. 
Moreover, with the increasing availability of different reporting modes on the same system, new degrees of freedom are introduced. Therefore, 
the utility of data fusion is also to achieve a more unified framework and a global vision of the system. On the other hand, there are issues and challenges  in doing that, as addressed by \cite{7214350}. Many choices can impact the results of data fusion such as the selected architecture, the customization of the methodology, and the technology tuning, which can lead to an optimal solution \cite{article44}.


Many studies have analyzed how to merge SAR and optical satellite data for Earth surface classification, discussing the results, possible improvements, and challenges. 
For instance, in  \cite{article63} a fusion of S-1 and S-2  satellite images for wetlands' classification is carried out. S-1 and S-2 datasets are created by downloading the data from the Copernicus Open Access Hub. 
Several methods of classification have been applied to  the Balikdami wetland in the Anatolian part of Turkey chosen as the study area.
The results show a significant improvement in the classification of wetlands when  both microwave and multispectral data are used.
In a  second article,  \cite{LUO20211944}, authors aim to map crop distribution in Heilongjiang Province using GEE  and S-1 and S-2 images in the period of crop growth, May to September 2018. Monthly composite images of reflectance bands, vegetation indexes, and polarization bands have been combined as input characteristics, by showing how better results are achieved in this case.
A further outcome highlights that the classification results using S-1 images in areas with large graphics are not significantly different from those using S-2 images. However, they differ a lot on small plots since S-1 data  do not make a proper distinction of batch boundaries, and the speckle noise affects results in this case.\\
In our study, the objective is to demonstrate the advantages of using combined SAR and optical data for the classification of a very fast-growing area in the USA: Texas City and its suburbs, making some consideration on the use of S-1 data. ML algorithms will be implemented on GEE and the validation process, resulting in better accuracy, will help us to confirm the starting hypothesis.
\section{Data sources and processing environment}
\noindent Satellite data from the \href{https://www.esa.int/Applications/Observing_the_Earth/Copernicus/Europe_s_Copernicus_programme}{European Space Agency (ESA) Copernicus mission} are considered for this study. Moreover, the environment of  \href{https://earthengine.google.com/}{GEE}  is utilized for downloading and processing the data. 
\subsection{Sentinel-1}
The European Radar Observatory for the Copernicus project, a joint venture between the European Commission and ESA, is known as the 
\href{https://sentinels.copernicus.eu/web/sentinel/missions/sentinel-1/overview}{S-1 mission}, 
operating in a pre-programmed, conflict-free mode, and capturing high-resolution images of the planet's landmasses, coastal regions, and ocean vignettes. S-1A  and S-1B represent a constellation of twin satellites, sharing the same orbital plane, phased at 180 degrees, although S-1B ended its operations in August 2022. S-1 satellites have four C-band imaging modes, with resolutions down to 5 meters and coverage ranges up to 400 km. The mission offers dual polarization, fast return times and product delivery. SAR technology allows for data collection across locations during day and night, regardless of weather conditions, and operates at wavelengths unaffected by cloud cover or light presence.
\subsection{Sentinel-2}
A wide-swath, high-resolution, multi-spectral imaging mission from Europe is represented by 
\href{https://sentinel.esa.int/web/sentinel/missions/sentinel-2/overview}{S-2}.
S-2A, which was launched on 23 June 2015, and S-2B, which was launched on 7 March 2017, are also twin satellites that fly in the same orbit  phased at 180 degrees. Their entire mission specification calls for a high revisit frequency of five days at the equator. Over the 13 available bands, four have a spatial resolution of 10 meters, six  a resolution of 20 meters, and three bands a resolution of 60 meters, sampled by the optical multi-spectral instrument (MSI) on board 
\subsection{Google Earth Engine}
The 
\href{https://earthengine.google.com/}{GEE platform} 
has been created to provide petabyte-scale geographic data display and scientific research. GEE offers a consolidated environment with extensive data collection, with diverse bands,  projections, and resolutions, enabling quick multisensor analysis.  It has built-in capabilities to allow users to build and use basic and advanced techniques, such as for instance ML-based algorithms and models, for analyzing Earth Observation (EO) data. Among the 
ML models, GEE makes at hand  common scenarios with easy-to-use Application Programming Interfaces (APIs). An important and very useful tool works to classify the pixels of satellite imagery into two or more categories. The approach is useful for Land Use Land Cover (LULC) mapping and other popular applications. There are different ML techniques that can be used, such as Supervised and Unsupervised Classification 
\href{https://developers.google.com/earth-engine/guides/machine-learning}{Supervised and Unsupervised Classification in GEE}.
\\
In particular, in our investigation, we made use of the  package handling supervised classification by traditional ML algorithms running in GEE. These classifiers include Classification and Regression Trees (CART), RandomForest, NaiveBayes, and Support Vector Machine (SVM). We chose the CART 
"ee.Classifier.smileCart()" tool. The CART algorithm was first published by Leo Breiman in 1984 \cite{Breiman}, and is a predictive model based on if-else decision rules. 
Training and test data for the ML models are created as "FeatureCollection", that is GeoJSON objects collection with geometry and properties.
To train the classifier, you need to specify the name of the class label  and a list of properties in the training table that the classifier must use for prediction. 
Once the classifier has been trained,  
\href{https://developers.google.com/earth-engine/guides/classification}{it is possible to proceed with classification and validation}.
In GEE, the validation step required by the classification process  is crucial, allowing to analyze classification accuracy.  \href{https://www.gears-lab.com/intro_rs_lab7/#classification-validation}{An error matrix} is created, useful for process analysis and for the accuracy calculation of individual classes.
\section{Area of Interest and ML model}
In our study, the Area of Interest (AOI) chosen as a benchmark for the analysis has been a very fast-growing area in the USA: Texas City and its
suburbs. Texas City is located in Galveston County, Texas, as shown in Figure \ref{city}, where the AOI coordinates are indicated on the caption.  

\begin{figure}[!ht]
    \centering
    \subfigure[]{\includegraphics[width=0.14\textwidth]{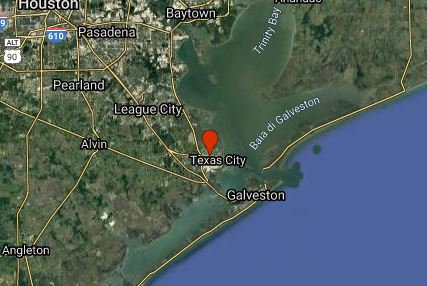}} 
    \subfigure[]{\includegraphics[width=0.13\textwidth]{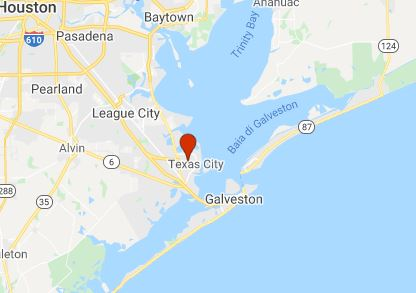}} 
    \caption{Area of interest: Texas City. 
    The coordinates of the point of interest are -94.925, 29.389. 
    }
    \label{city}
\end{figure}

\section{Dataset Creation and Methodology}
The data used and the methodology developed for the purposes of this case study will be analyzed below.
\subsection{Methodology for Classification}
Starting from the Texas City coordinates,  
it was possible to select and collect a series of  S-2 images  of the AOI. One of them  is represented in Figure \ref{Sentinel2image} as an example. 
We  established a period of analysis from the first of January 2020 to the first of August 2021, and  
we selected  the  B2-B3-B4-B5-B6-B7 bands of the images with less cloud coverage. 
After dataset preparation, the CART classifier, chosen in our case as the Supervised ML algorithm in GEE, has been settled and trained. \\
\begin{figure}[ht!]
	\centering
	\includegraphics[scale=0.21]{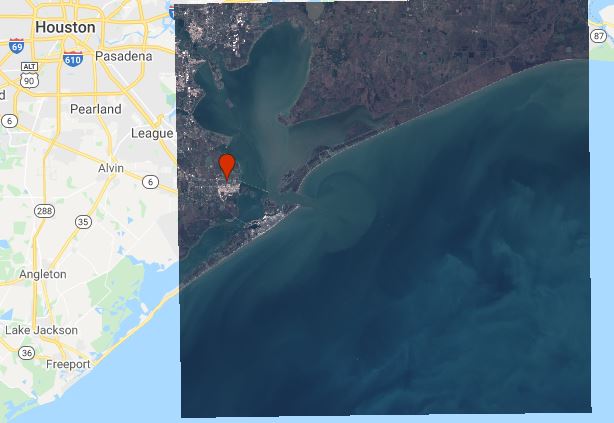}
	\caption{ Example of a  Sentinel-2 image over the AOI}
	\label{Sentinel2image}
\end{figure}
In particular, we chose to analyze three classes:
\begin{itemize}
    \item water area;
    \item urban area;
    \item non-urban area.
\end{itemize}
For each class, the number of samples used for training is reported in Table \ref{tabellaclass}. It is worth to highlight that these elements were taken manually on the image of interest through visual inspection.

\begin{table}[ht!]
    \centering
    \begin{tabular}{|l|c|r|}
     \hline
     Classes & Number of samples  \\
     \hline
     Water    & 78 \\
     \hline
     Urban   &53 \\
     \hline
     Non-urban   &70\\
     \hline
\end{tabular}
\vspace{2pt}
    \caption{For each class, the table reports the number of samples used for the CART classifier training. }
    \label{tabellaclass}
\end{table}

Therefore, the first classification of the AOI was carried out based on S-2 images.
After that, also S-1 data have been retrieved over the same time period. SAR bands (i.e. different available polarizations, Vertical-Vertical (VV), and Vertical-Horizontal (VH)) have been integrated with the optical bands  to create a composite image on which to proceed with a new classification. \\In Figure  \ref{workflow} the general workflow is represented. To obtain 
\begin{figure}[ht!]
	\centering
\includegraphics[scale=0.50]{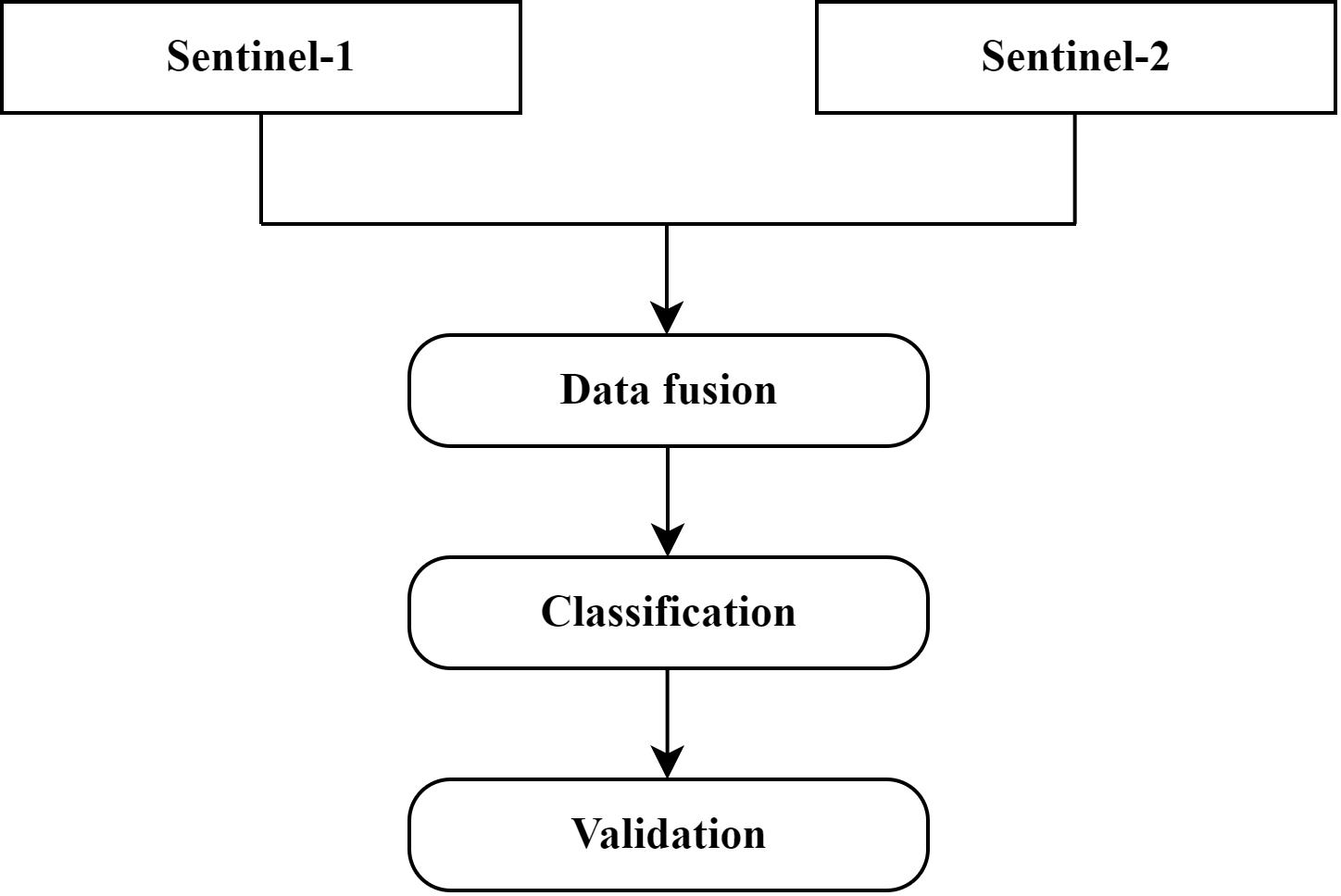}
	\caption{Classification with Sentinel-1 and Sentinel-2 data}
	\label{workflow}
\end{figure}
the fusion of different images  some operations have been performed, as specified ahead,  to make  it possible for the classifier to use  both optical and SAR data. \\First, imported S-1 data are  spatially and temporally filtered. The next step is to create a stack of S-1 images, a multiband multidimensional new image. In this case, it is a 4-band image, incorporating both the viewing angles and the VV and VH polarizations, so we are including as much information as possible for the final classification.
Moreover, each band is the result of reducing over the period of interest the data to the average value of each pixel, which is a very effective way to eliminate or strongly reduced radar speckle noise.
\subsection{Methodology for Validation}
The objective of this analysis, as mentioned above, was to demonstrate
that the integration of optical and SAR data could improve classification accuracy. 
The number of samples used for each class for the validation step is reported in Table \ref{tabellavalidazione}, where also in this case the class samples were taken manually on the image of interest. The values of accuracy are derived from the confusion matrixes, showing that S-2 data alone allow to achieve a precision of
81.7\%, while the integration of optical and SAR data
allow it to reach  88.8\% as explained in the next section.
\begin{table}[ht!]
    \centering
    \begin{tabular}{|l|c|r|}
     \hline
     Classes & Number of samples  \\
     \hline
     Water    & 129 \\
     \hline
     Urban   &95 \\
     \hline
     Non-urban   &89\\
     \hline
\end{tabular}
\vspace{2pt}
    \caption{Number of samples for each class for the validation process}
    \label{tabellavalidazione}
\end{table}
\section{Results}
\subsection{Images for Classification}
In this section, the images prepared for classification and validation are shown, and results are presented and discussed. 
\begin{figure}[ht!]
	\centering
     \includegraphics[scale=0.24]{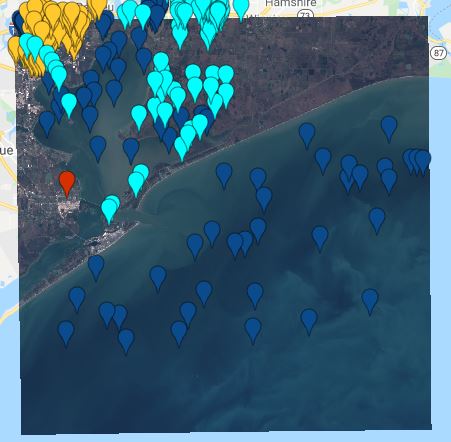}
	\caption{Classes over the image used for training the CART classifier.}
	\label{clastraining}
\end{figure}
Figure \ref{clastraining} shows the image used to train the CART classifier. In particular, several pins have been placed for identifying each class:  the light blue pins represent the training elements for the non-urban area class; the yellow pins represent the training items of the urban class and finally, the blue pins represent the training elements for the water area class. The red pin identifies Texac City's position.\\
Specific zoomed zones of the whole region of interest were taken into consideration to better show the  results obtained from this study. In particular, from Figures \ref{fig5} and \ref{fig6}, we can get preliminary qualitative 
 results of the proposed method. Account should be taken of the fact that the red zones identify non-urban areas; those in white identify urban areas and finally, the blue zones identify water areas. 

\begin{figure*}[!ht]
    \centering
    \subfigure[]{\includegraphics[width=0.20\textwidth]{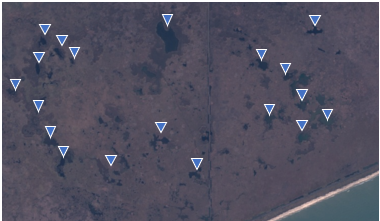}} 
    \subfigure[]{\includegraphics[width=0.2039\textwidth]{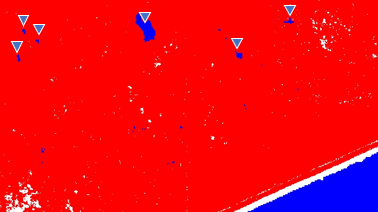}} 
    \subfigure[]{\includegraphics[width=0.201\textwidth]{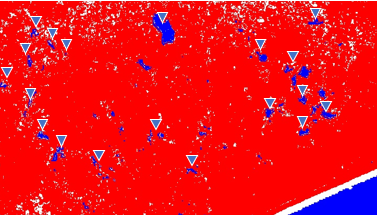}} 
    \caption{(a) represents the image used for training the CART classifier, (b) represents classification when only S-2 images are considered, (c) represents classification performed on the composite image by joining optical bands with SAR bands from Sentinel-2 and Sentinel-1 data. 
    }
    \label{fig5}
\end{figure*}
\begin{figure*}[!ht]
    \centering
    \subfigure[]{\includegraphics[width=0.201\textwidth]{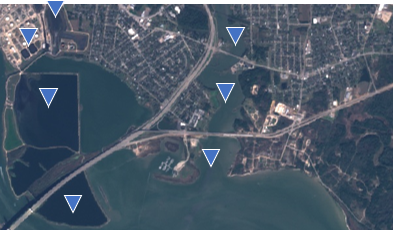}} 
    \subfigure[]{\includegraphics[width=0.20\textwidth]{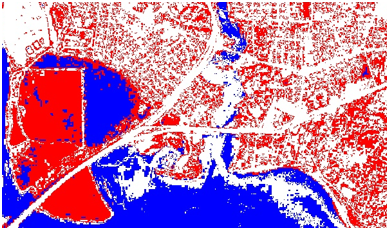}} 
    \subfigure[]{\includegraphics[width=0.204\textwidth]{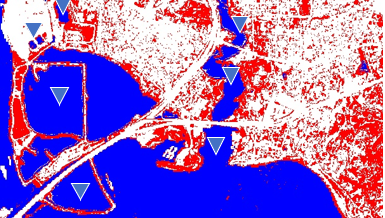}} 
    \caption{(a) represents the image used for training the CART classifier, (b) represents classification when only S-2 images are considered, (c) represents classification performed on the composite image by joining optical bands with SAR bands from Sentinel-2 and Sentinel-1 data.
    }
    \label{fig6}
\end{figure*}
In particular, from Figure \ref{fig5}, (a) represents the image used for training the CART classifier,  (b) represents classification when only S-2 images are considered, and  (c) represents classification performed on the composite image by joining Sentinel-2 optical bands with Sentinel-1 SAR bands. In this latter case,  classification is performed on the composite image acquiring information dictated by the sum of the optical and radar bands, the latter including viewing angles and polarizations. Geometrical information is added to the scene of interest, bringing better results. In fact, as far as the qualitative comparison between the images is concerned, it is possible to notice that image (b) cannot keep track of all the regions of water while image (c) takes into account most of the water regions, signed with blue color indicators that have been positioned above those regions to better highlight their characteristics. This wants to give a first example of how the classification could benefit from composite images.\\
Another example is shown in Figure \ref{fig6}, where (a) represents the image used for training the CART classifier in a different area, (b) represents classification when only S-2 images are considered, and  (c) represents classification performed on the composite image by joining optical and SAR bands.
As can be seen from this image, the classification is more accurate on the composite image. The blue indicators represent the water areas and they are shown in the first image used for training the CART classifier. When the classification is performed with only optical data, it is noticed that the water areas have not been detected correctly, on the contrary referring to the areas on the left side of the image for instance, they have been classified as non-urban areas since they appear in red color. It can also be noticed that the watercourse in the original image is not correctly identified as a water zone in the image with only optical data, in fact,  it is classified as an urban area with the color white. All these issues appear solved in case (c).

\subsection{Images for Validation}
For the validation, the number of samples taken into consideration for each class  has been higher, as reported in Table \ref{tabellavalidazione}. The several pins allocated over the image are represented in Figure \ref{validation}, distinguished  for water area, urban area, and non-urban area. 
\begin{figure}[ht!]
	\centering
   \includegraphics[scale=0.26]{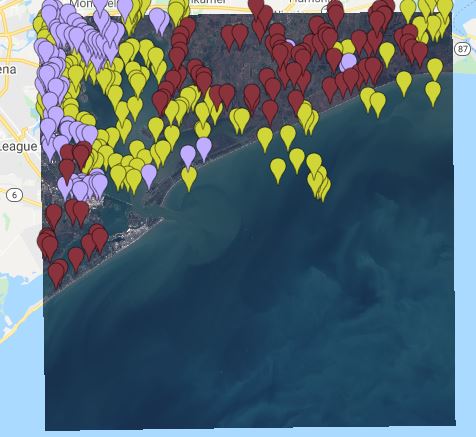}
	\caption{Classes over the images for the validation process on the area of interest (AOI).}
	\label{validation}
\end{figure}

Quantitative results have also been obtained and analyzed. Accuracy with S-2 data only  reaches 81.7\%, while if  SAR data are integrated as discussed before, final accuracy reaches 88.8\%, by 
proving our initial hypothesis of reaching better performance for the classification of an AOI with SAR and optical data integration. 
\section{Conclusions}
Our work demonstrated the advantages of integrating optical and SAR data for Earth's surface classification when a partic\nobreak ular stack of SAR images is considered, and when supervised ML algorithms implemented on GEE are utilized. Qualitative and quantitative results showed that   a final better accuracy is reached  for the classification of the Texas City area chosen as the case study. The extension of this analysis is in progress, aiming to compare supervised and unsupervised methods and to include further speckle filtering pre-processing on SAR data.
\vspace{-0.4cm}
\bibliographystyle{IEEEtran}
\bibliography{ref}
\end{document}